\documentclass[10pt, a4paper]{article}

\usepackage{lrec-coling2024} 
\usepackage{multirow}
\usepackage{booktabs}
\usepackage{graphicx}
\usepackage{url}
\usepackage{bbding}
\usepackage{tabularx}
\usepackage{hyperref}

\newcommand{\M}{CMNEE}

\title{ \vspace*{.5\baselineskip}  \textbf{\M{}: A Large-Scale Document-Level Event Extraction Dataset based on Open-Source Chinese Military News}}

\name{
  \begin{tabular}{c}
    Mengna Zhu$^{1}$, Zijie Xu$^{2}$, Kaisheng Zeng$^{3}$, Kaiming Xiao$^{1}$, Mao Wang$^{1}$, \\
    Wenjun Ke$^{2}$, Hongbin Huang$^{1}$\sthanks{\ \ Corresponding author}\\
  \end{tabular}
}

\address{
    $^{1}$ Laboratory for Big Data and Decision, National University of Defense Technology\\
    $^{2}$ School of Computer Science and Engineering, Southeast University \\
    $^{3}$ Computer Science and Technology, Tsinghua University \\
    \texttt{
        \begin{tabular}{c}
            \{zhumengna16, kmxiao, wangmao, hbhuang\}@nudt.edu.cn\\
            \{zijiexu, kewenjun\}@seu.edu.cn\\
            \{zks19\}@mails.tsinghua.edu.cn\\
        \end{tabular}
    }
    }

\abstract{
Extracting structured event knowledge, including event triggers and corresponding arguments, from military texts is fundamental to many applications, such as intelligence analysis and decision assistance.
However, event extraction in the military field faces the data scarcity problem, which impedes the research of event extraction models in this domain.
To alleviate this problem, we propose \textbf{\M{}}, a large-scale, document-level open-source \textbf{C}hinese \textbf{M}ilitary \textbf{N}ews \textbf{E}vent \textbf{E}xtraction dataset. 
It contains 17,000 documents and 29,223 events, which are all manually annotated based on a pre-defined schema for the military domain including 8 event types and 11 argument role types. We designed a two-stage, multi-turns annotation strategy to ensure the quality of \M{} and reproduced several state-of-the-art event extraction models with a systematic evaluation. The experimental results on \M{} fall shorter than those on other domain datasets obviously, which demonstrates that event extraction for military domain poses unique challenges and requires further research efforts. Our code and data can be obtained from \url{https://github.com/Mzzzhu/CMNEE}.
 \\ \newline  }

\begin{document}

\maketitleabstract

\section{Introduction}
Event extraction refers to the extraction of structured information from unstructured text, which is typically separated into two subtasks: event detection and event argument extraction~\citep{ahn-2006-stages,yang-etal-2019-exploring,xu-etal-2023-learning-friend,wang-etal-2021-cleve,shi-etal-2023-hybrid,yang-etal-2023-amr,wan-etal-2023-joint,liu-etal-2020-event, cdeee, peng2023omnievent, wang2023mavenarg}. Current research predominantly focuses on the general news or financial domains, with only relatively fewer studies for military domain. Nevertheless, the importance of military event extraction is undeniable.

\begin{figure*}[!ht]
\begin{center}
\includegraphics[scale=0.45]{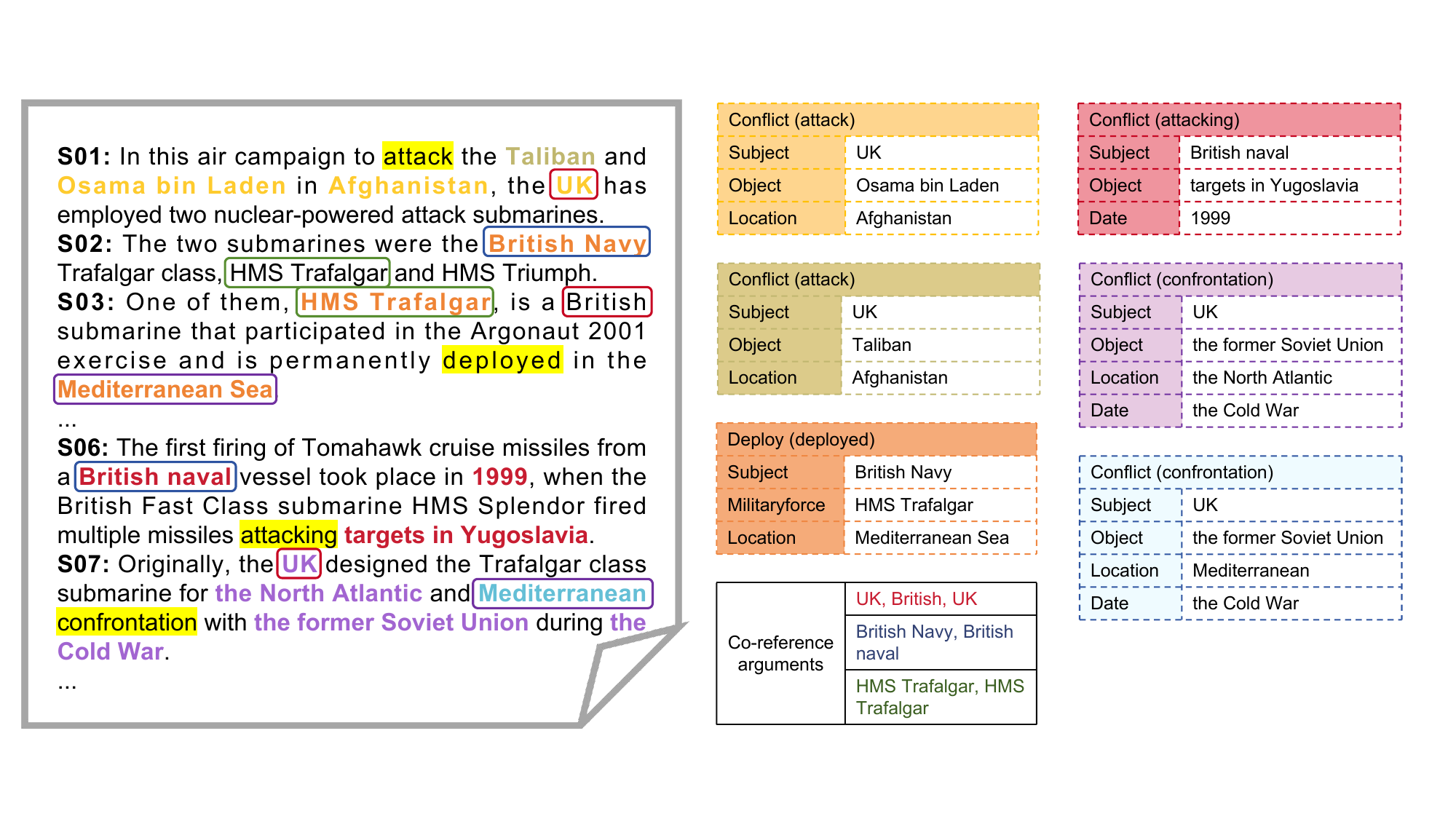} 
\caption{An example from \M{}. Each document in \M{} is annotated with event types and involved event arguments. This document contains 6 events. We used different colors to distinguish arguments of these events and colored rectangular boxes to mark different co-reference arguments.}
\label{fig.dee-sample}
\end{center}
\end{figure*}

The military domain encompasses numerous documents which contain rich event information. Extraction of military events from documents is crucial to downstream applications such as intelligence analysis~\citep{Julia,Martin,Lawrence,Ivanov}, decision making assistance~\citep{skryabina2020role}, and strategic planning~\citep{schrodt2012precedents,RVallikannu,Hui}. Figure~\ref{fig.dee-sample} provides an example of event extraction based on military news. This military news document contains six events. It's necessary to recognize triggers in the document and determine their corresponding event types, as well as relevant arguments and corresponding roles. For instance, in the 6th sentence, S06, it's imperative to recognize the trigger \emph{attacking} and the event type \emph{Conflict}. Additionally, the ``Subject'' \emph{British Naval}, the ``Object'' \emph{targets in Yugoslavia}, and ``Date'' \emph{1999} of the event should also be identified.

Existing research on event extraction largely relies on deep neural network-based models, which have a strong dependence on data size and quality. Consequently, the lack of high-quality training data significantly limits their performances. Most of these models use sentence-level text for analysis, and while sentence-level event extraction research is relatively mature ~\citep{Gaohongbo,DEGREE,HUANG}, this technology struggles when directly applied to document-level event extraction tasks. This is due to with the increased length of the text, event arguments often scattered across different sentences.
Moreover, most existing event extraction datasets are oriented towards general~\citep{DUEE,WIKIEVENTS} or financial domains ~\citep{duee-fin, doc2edag}. The performance of methods applied to datasets in different languages and domains often varies significantly. Currently, military event data extraction primarily relies on human labor, leading to issues such as low efficiency, inconsistent standards, and incomplete information. These challenges make it difficult to support practical applications based on events. Therefore, constructing a document-level event extraction dataset specifically for the military domain holds substantial practical significance and application value.

However, due to the confidentiality and sensitivity of military domain data and the difficulty of obtaining event schema, there is a long-term gap in the datasets used for the event extraction task in this domain, which is the main reason that the development of military event extraction is seriously lagging behind mainstream research. 

To fill this gap and alleviate relevant problems, we proposed a large-scale document-level open-source \textbf{C}hinese \textbf{M}ilitary \textbf{N}ews \textbf{E}vent \textbf{E}xtraction dataset (\textbf{\M{}}), which involved corpus from authoritative websites such as Huanqiu\footnote{https://www.huanqiu.com/}, China Military Online\footnote{http://www.81.cn/wzsy/index.html}, Sina Military\footnote{https://mil.news.sina.com.cn/} and Baidu Encyclopedia\footnote{https://baike.baidu.com/}.
The acquired military news text itself is reliable. In the process of data annotation based on the news, firstly, the event types that may be contained in the document are pre-labeled by matching with constructed triggers dictionary, combined with domain experts' knowledge and the analysis of related news to form the event schema. Based on the event schema and referring to the data annotation criteria, the annotation is completed in a two-stage multi-turns manner. The annotation requirements are to make manual judgments on the results of pre-labelling, annotate the triggers and event types, event arguments and corresponding argument roles. In addition, considering that one of the reasons for the poor performance of event argument extraction is that the identified arguments are co-reference arguments rather than those labeled in the dataset~\citep{lu2023event}, the co-reference arguments involved in the text are also labeled. Quality indicators are designed to evaluate \M{}. Annotations and data annotation criteria are continuously corrected until all indicators reach the specified thresholds.

\M{} is currently the only dataset for the document-level event extraction task in the military domain. To better evaluate \M{}, we implemented several advanced models and conducted a systematic evaluation. The experimental results demonstrate that \M{} has some unique challenges that need to be overcome and deserves more efforts to study. We hope that \M{} can facilitate relevant research, and attract more attention into event extraction for military domain.

\section{Construction of \M{}}
\begin{figure*}[!ht]
\begin{center}
\includegraphics[scale=0.45, trim = 10 150 5 145, clip]{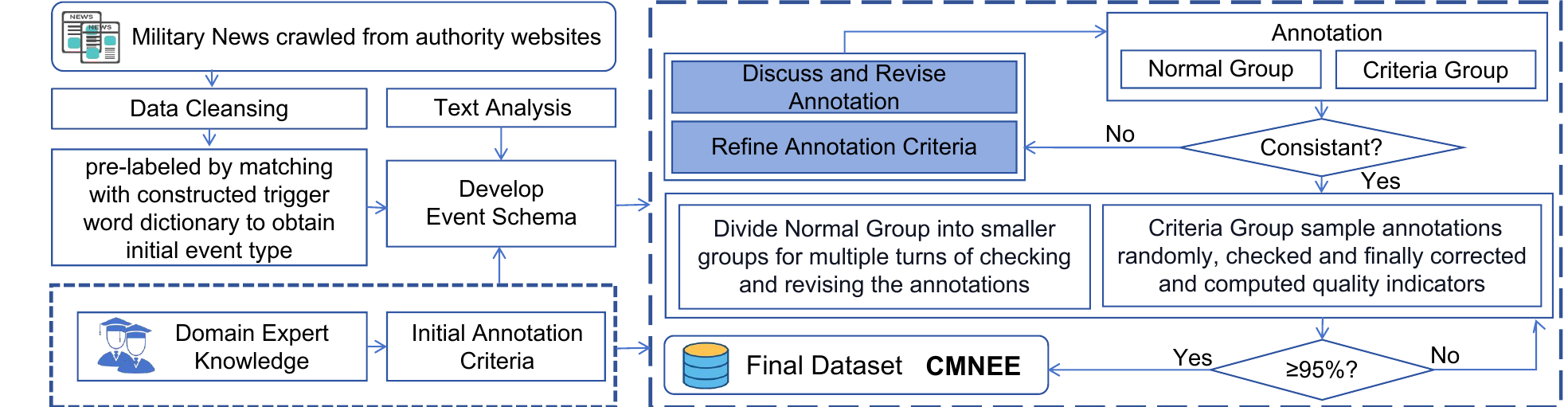} 
\caption{Main process of \M{} construction. Firstly, crawled military news texts from websites with authority and pre-processed data. Then manually annotated them in groups after completing pre-labeling. Annotation is constantly corrected by using a two-stage multi-turns annotation approach until all the indicators reached the specified thresholds.}
\label{fig.annotation-flow}
\end{center}
\end{figure*}

Our main goal is to construct a large-scale dataset that facilitates the development of event extraction in the military domain. During the construction of \M{}, event trigger, event type, event arguments, argument roles and co-reference arguments need to be labelled. The main process of \M{} construction is shown in Figure \ref{fig.annotation-flow}. Firstly, we crawled military news texts from websites with authority and then pre-processed data with the help of existing mature tools. Then manually annotated them in groups after completing automatic event types pre-labeling based on predefined candidate triggers dictionary. Annotation is constantly corrected by using a two-stage multi-turns annotation approach until all the quality indicators reached the specified thresholds. We finally obtained 17,000 pieces of high-quality labelled documents data.

\subsection{Data Collection and Preprocessing}
To ensure the quality of the dataset, our data sources are military news websites with authority, relevance, and domain value, such as Huanqiu, China Military, Sina Military, and Baidu Encyclopedia. We used Python's Requests library and BeautifulSoup library to crawl nearly 100,000 news data from above websites. Based on the 7 predefined event types in the sentence-level military event detection dataset MNEE~\citep{hongbin-2023}, we used ChatGPT\footnote{https://chat.openai.com/} to generate synonyms and related words for these event types, which were manually filtered to form a keyword dictionary, and used words in the dictionary to match with news text, according to the number of matching words to filter out 30,000 pieces of data with high relevance. Then the Python Pandas library was used to filter the missing samples and remove the duplicated samples, and regular expressions were used to remove the garbled codes and the meaningless information such as the html of the web pages. Finally 17000 high quality news documents were retained.

\subsection{Event Schema Construction and Pre-labeling}
Following MNEE, combining the analysis of collected news and referring to domain experts' knowledge, 8 event types other than non-event are defined, including Experiment, Manoeuvre, Deploy, Support, Accident, Exhibit, Conflict, and Injure. We constructed a trigger dictionary and pre-labeled event types that each sample may contain, and used the pre-labeled results to conduct preliminary statistical analysis to obtain event type proportion, and referred to domain experts' knowledge in combination with the real situation to assess the reasonableness of the distribution of event types.

Then, event argument roles are determined by analyzing documents corresponding to each event type obtained after pre-labelling. The final event schema can be seen in Table \ref{Event-schema}.

\begin{table*}
\centering
\begin{tabular}{lp{7.6cm}l}
\toprule
\textbf{Event type} & \textbf{Definition} &  \textbf{Argument Roles}\\
\midrule
Experiment & Countries, organizations, institutions, or companies verify whether the equipment is qualified based on existing standards, including but not limited to flights, launching, testing, etc. & Subject, Equipment, Date, Location \\
Manoeuvre & Exercises conducted in the course of campaigns and battles, in the context of the situation envisaged, and are the highest and most centralized form of military training. & Subject, Content, Date, Area \\
Deploy & The movement of military forces, i.e., personnel or equipment, within a country or organization. & Subject, Militaryforce, Date, Location\\
Support & Subject provides material help or relief actions to the object. &  Subject, Object, Materials, Date\\
Accident & Accidents that happen unexpectedly. & Subject, Result, Date, Location \\
Exhibit & Countries, organizations, institutions, companies, etc. exhibit or publicize equipment, products, etc. through airshows or other forms. &  Subject, Equipment, Date, Location \\
Conflict & Violence acts, such as attack, that causes damage or injury, or a conflict or confrontation between two parties, such as protest or condemnation. & Subject, Object, Date, Location \\
Injure & Person entity suffered physical injury. & Subject, Quantity, Date, Location \\
\bottomrule
\end{tabular}
\caption{\label{Event-schema} Event schema of \M{}}
\end{table*}

\subsection{Human Annotation and Quality Evaluation}
After completing data preprocessing, the pre-labeled event type information is used as a reference to assist human annotation and labelers need to judge them first of all. In the process of data labelling, two-stage multi-turns iterative approach is adopted to continuously revise the annotation so as to enhance the quality of the dataset.

\textbf{Stage 1}: In order to improve the efficiency of labelling while keeping the quality as high as possible, we divide labelers into two groups, 1) Normal Group, members of which are employees of the labelling company engaged in the labelling work for a long time, with partial background knowledge. and 2) Criteria Group, members of which are domain experts with relatively complete professional knowledge and involved in the formulation of data annotation criteria. All the data are shuffled and processed in batches, personnel of these two groups respectively labelled documents by batch at the same time. When the acquired documents have been labelled, the annotation should be checked, and labels of disagreement should be discussed and amended in time. Meanwhile, data annotation criteria is continuously refined.

\textbf{Stage 2}: The Normal Group was divided into several smaller groups to examine annotation results of Stage 1 in multiple turns by group separately, and labels were corrected according to the final data annotation criteria. At the same time, Criteria Groups randomly select samples to examine and discuss corrections within the group.

In order to assess the quality of the dataset, four types of quality indicators, namely, Event Type Accuracy, Event Type Recall, Event Argument Recall, and Co-reference Argument Recall, are used with reference to MNEE \citep{hongbin-2023}, definitions can be seen in Table~\ref{quality-indicators}.  Detailed descriptions are as follows. $Cor_{es}$ represents the number of samples whose events are all correctly labelled. $S$ represents the total number of selected samples. $Cor_e$, $Cor_a$, $Cor_c$ represent the number of correctly labelled events, arguments, co-reference arguments in the sample respectively. $Act_e$, $Act_a$, $Act_c$ represent the number of actually existed events, arguments, co-reference arguments in the sample respectively. Correctly labelled events refer to events whose triggers with their offsets and corresponding event types are all correctly labelled. Correctly labelled event arguments refer to arguments that have the correct texts, roles, and offsets under the condition that their event types are correct. Actual events, arguments, co-reference arguments refer to those that are expected to be labelled in accordance with the data annotation criteria.

\begin{table*}
\centering
\setlength{\tabcolsep}{3.5mm}{
\begin{tabular}{lll}
\toprule
\textbf{Indicators} & \textbf{Abbreviations} & \textbf{Calculation Methods}\\
\midrule
Event Type Accuracy & $TypeAcc$ & $(Cor_{es} / S) \times 100\% $  \\
Event Type Recall & $TypeR$ & $(Cor_e / Act_e) \times 100\%$ \\
Event Argument Recall & $ArgR$ & $ (Cor_a / Act_a) \times 100\%$ \\
Co-reference Argument Recall & $Co-refR$ & $(Cor_c / Act_c) \times 100\%$ \\
\bottomrule
\end{tabular}
}
\caption{\label{quality-indicators}Definitions of Quality Indicators}
\end{table*}

After annotation is completed, the Criteria Group checks a certain percentage of randomly selected samples many turns. If all indicators of the samples are higher than 95\%, the dataset is qualified, or else the Normal Group will continue to correct the annotations. Final quality indicator results are 95.1\%, 96.4\%, 96.8\%, and 97.2\% respectively.

\section{Data Analysis of \M{}}
\begin{figure}[!ht]
\begin{center}
\includegraphics[scale=0.42,trim=200 170 50 100,clip]{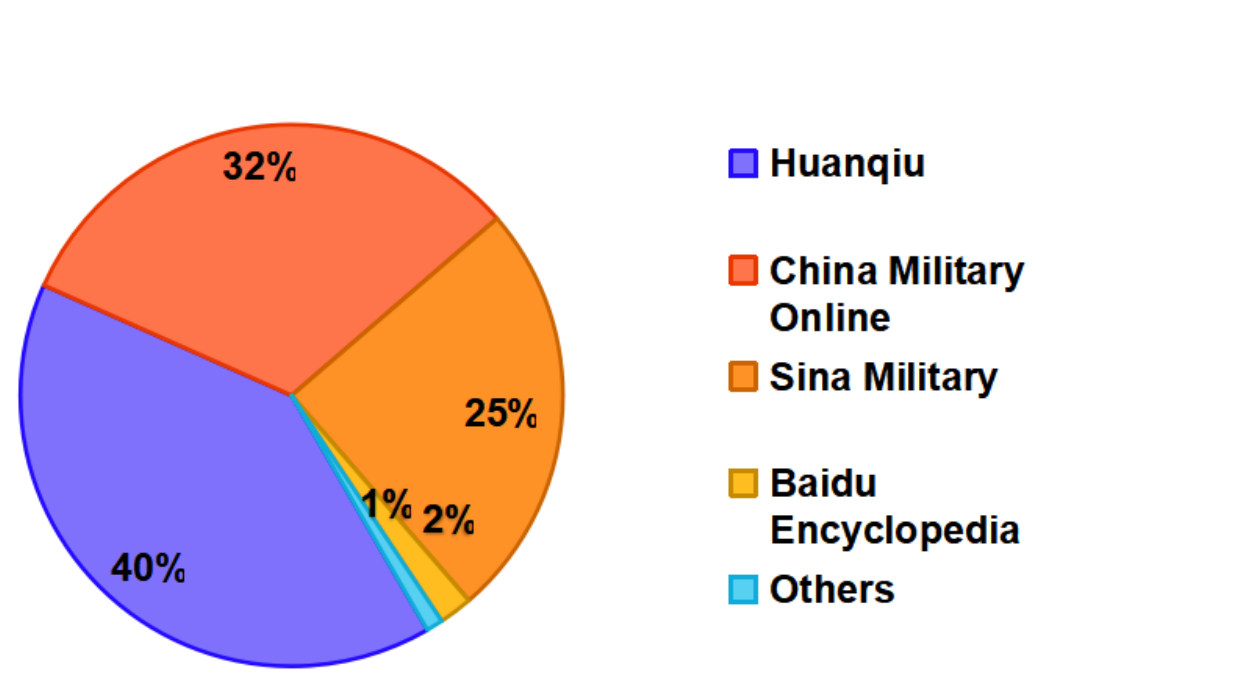} 
\caption{Data Sources}
\label{fig.data-sources}
\end{center}
\end{figure}

In this section, we analyze various aspects of \M{} to provide a deep understanding of the dataset. Overall, \M{} defines 8 event types and 11 argument roles. It contains 17,000 valid documents sourced from authoritative websites, 29,223 non-empty events and 93,348 event arguments. The distribution of data sources is shown in Figure \ref{fig.data-sources}, which shows that the vast majority of documents are from reliable sources. On average, there are 330 tokens, 6.7 sentences, 1.8 events and 6.6 event arguments per document in \M{}. The longest number of document tokens in the \M{} is 496 and the largest number of sentences is 17. Tokens here refer to Chinese characters.

We compare \M{} with some existing widely-used EE related datasets and show the main statistics in Table \ref{compared-with-other-datasets}, including sentence-level event extraction datasets, ACE 2005~\citep{ace2005}, MAVEN ~\citep{MAVEN}, DuEE ~\citep{DUEE}, MNEE ~\citep{hongbin-2023} and document-level event extraction datasets, RAMS~\citep{ebner-etal-2020-multi}, WikiEvents~\citep{li-etal-2021-document}, Duee-fin~\citep{dueefin}, ChfinAnn~\citep{doc2edag}, DocEE~\citep{tong-etal-2022-docee}. \M{} is comparable with existing datasets in the number of documents and events. Despite its relatively small number of event types and argument roles, \M{} basically covers the main event types in the military domain.

\begin{table*}
\centering
\setlength{\tabcolsep}{1.5mm}{
\begin{tabular}{llllrrrrcc}
\toprule
{\textbf{Level}} & \textbf{Dataset} &  \textbf{Domain} & \textbf{Language} & \textbf{Docs} & \textbf{EvTyps} & \textbf{ArgRs} & \textbf{Events} & \textbf{ED} & \textbf{EAE} \\
\midrule
\multirow{4}{*}{\textbf{Sentence}} & ACE2005 & General & English & 599 & 33 & 35 & 4,090 & \Checkmark  & \Checkmark \\
~ & MAVEN & General & English  & 4,480 & 168 & - & 111,611  & \Checkmark & \XSolidBrush \\
~ & DuEE & Financial & Chinese & 11,224 & 65 & 121 & 19,640 & \Checkmark & \Checkmark \\
~ & MNEE & Military & Chinese &  13,000 & 8 & 10 & 6,997 & \Checkmark & \Checkmark \\
\midrule
\multirow{6}{*}{\textbf{Docment}} & RAMS & General & English & 3,993 & 139 & 65 & 9,124 & \Checkmark & \Checkmark\\
~ & WikiEvents & General & English & 246 & 50 & 59 & 3,951 & \Checkmark & \Checkmark \\
~ & Duee-fin & Financial & Chinese & 11,700 & 13 & 92 & 11,031 & \Checkmark & \Checkmark \\
~ & ChfinAnn & Financial & Chinese & 32,040 & 5 & 24 & 47,824 & \XSolidBrush & \Checkmark \\
~ & DocEE & General & English & 27,485 & 59 & 356 & 27,485 & \Checkmark & \Checkmark \\
~ & \textbf{\M{}} & Military & Chinese & 17,000 & 8 & 11 & 29,223 & \Checkmark & \Checkmark \\
\bottomrule
\end{tabular}
}
\caption{\label{compared-with-other-datasets}Statistics of \M{} compared with existing widely-used EE datasets}
\end{table*}

\subsection{Event Types Distribution}
\begin{figure}[!ht]
\begin{center}
\includegraphics[scale=0.045,trim=10 5 15 8,clip]{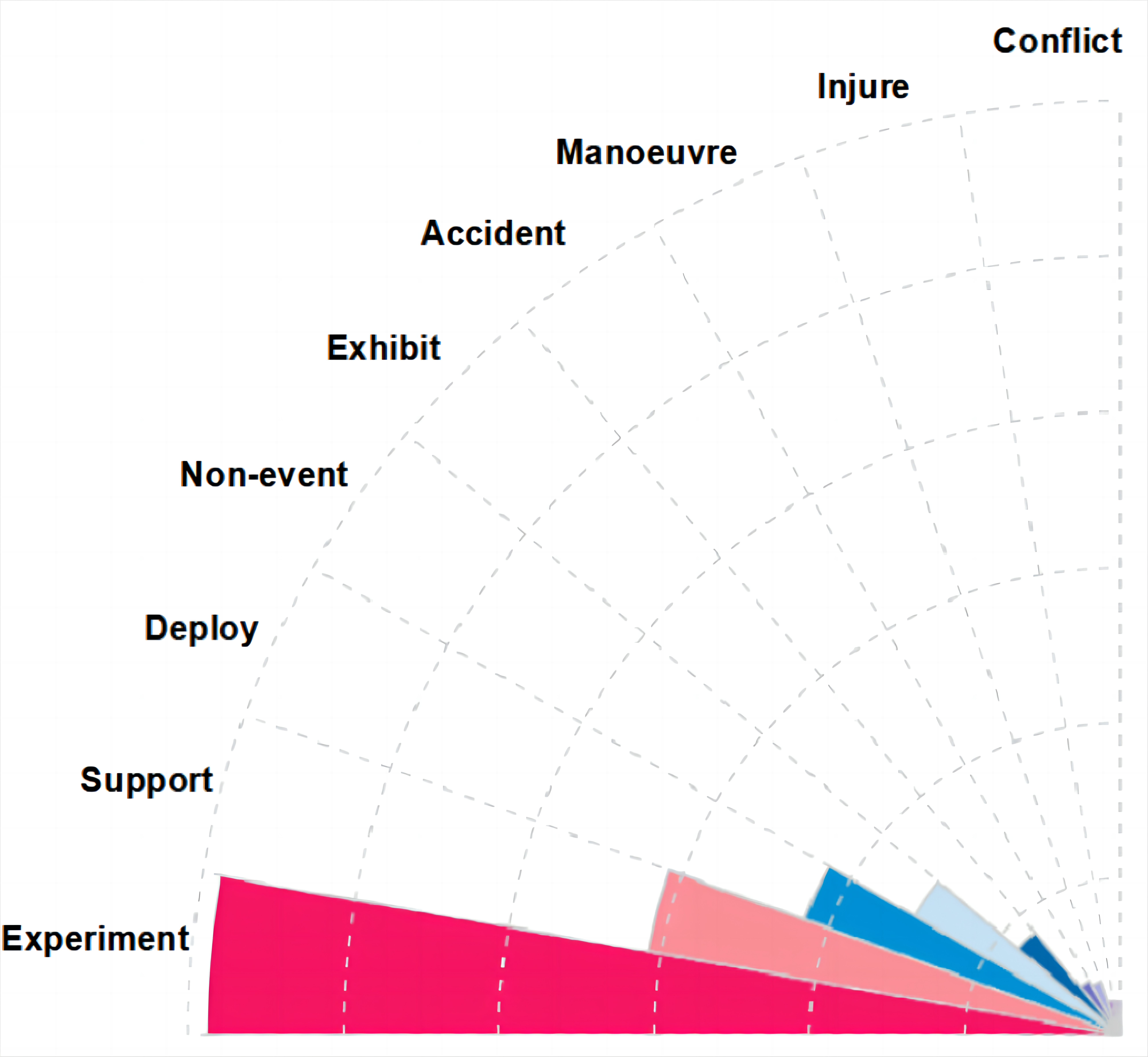} 
\caption{Event Type Distribution}
\label{fig.event-types}
\end{center}
\end{figure}

 Figure \ref{fig.event-types} shows the event types distribution of \M{} by their instance numbers. We can observe that the inherent data imbalance problem also exists in \M{}. It is obvious that \M{} is long-tailed and consistent with the characteristics of the data in the actual scenario. It is beneficial to use the models trained based on \M{} for developing real-world applications. Although \M{} is long-tailed, as \M{} is large-scale and event types are not particularly numerous, all event types have more than nearly 500 instances respectively. Conflict has the lowest number of instances as 477. The ``Experiment'' event type has the largest number of instances in the dataset, which is 13,027. \M{} is relatively less affected by data sparsity.

\subsection{Events Analysis}
\begin{figure}[!ht]
\begin{center}
\includegraphics[scale=0.045,trim=10 48 10 32,clip]{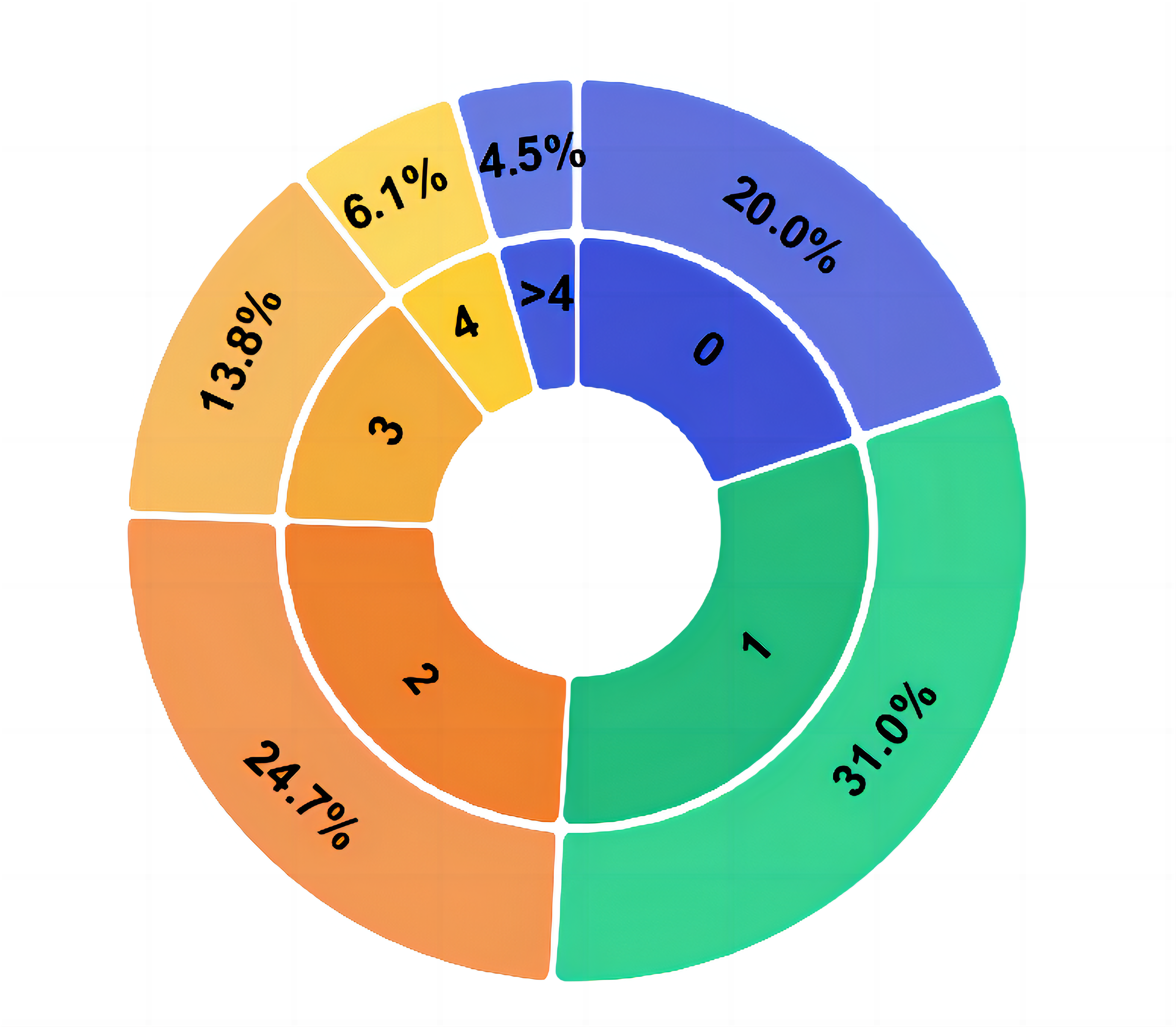} 
\caption{Multi-events Distribution}
\label{fig.multi-evens}
\end{center}
\end{figure}

Since the actual news may usually contain multiple events, during the annotation process, we try to annotate all the events information in the document as much as possible. Figure \ref{fig.multi-evens} displays the distribution of the number of events contained in each instance within the dataset. One instance corresponds to a single document. The inner circle numbers in the donut chart represent the quantity of events, while the outer circle numbers indicate the respective proportions. For example, the green segment denotes the proportion of documents that contain only one event. From this figure, it is evident that nearly half of the documents include two or more events, and approximately a quarter contain three or more events. So it is hopeful that \M{} will facilitate the research on multi-event extraction and promote the extraction more independently, completely and accurately.

In addition, due to the concise and condensed nature of military texts, there are a large number of overlapping events in CMNEE, and the percentage of instances containing overlapping events is 42\%. Overlapping events mean that the same word acts as a trigger for different events in the same instance, the same argument plays different roles in multiple events or the same argument plays different roles in the same event~\citep{sheng-etal-2021-casee}. For example, in the sentence ``The submarine was delivered on January 9 after completing trials.'', the ``submarine'' is both the material delivered and the equipment for the trials. Currently, the FewFC dataset is the main dataset used for the study of overlapping events~\citep{sheng-etal-2021-casee, cao-etal-2022-oneee}, but the size of this dataset is relatively small, in which only 18\% of the instances contain overlapping events, and its annotation is based on sentences. Overlapping events easily lead to confusion and omission of event information, which increases the difficulty of extracting event information completely.

\subsection{Event Arguments Analysis}
Since military text involves many proper nouns such as weapons and equipment, the proportion of long arguments is relatively high, and the proportion of arguments with more than 10 Chinese characters is nearly 17\%, which increases the difficulty of argument extraction.

Considering when there are multiple events in a document, it is likely that the same argument belongs to more than one event at the same time, so we defined \textbf{\emph{shared arguments}} as arguments that belong to different events at the same time with the same text or in the same co-reference arguments sub-list for ease of analysis. There are 19,353 shared arguments contained in \M{}, which is about one-fifth of the common arguments. Shared arguments can reveal relationships between events. For example, if two events share a \emph{Date} argument, it often indicates that these events occurred simultaneously. Accurately extracting all events which shared arguments based on \M{} is difficult, but it is very important. So more attention should be paid into it. It is also an effective way to improve the performance.

\section{Experiments on \M{}}
\subsection{Experimental setting}
\paragraph{Benchmark setting} Greedy algorithm is used to split \M{} into training, development, and test sets to make event types of these subsets as identically distributed as possible. Table \ref{split-\M} lists basic statistics of the split result.

\begin{table}
\centering
\setlength{\tabcolsep}{3.5mm}{
\begin{tabular}{lrrr}
\toprule
\textbf{Subset} & \textbf{Docs} & \textbf{Events} & \textbf{Args} \\
\midrule
Train & 12,000 & 19,427 & 62,591 \\
Dev & 2,000 & 3,236 & 10,275\\
Test & 3,000 & 6,560 & 20,842\\
\bottomrule
\end{tabular}
}
\caption{\label{split-\M}Statistics of splitting \M{}}
\end{table}

\paragraph{Models} We selected several advanced baselines for evaluation and they can be categorized into three types. The first category includes models that are designed based on Chinese corpora and do not use triggers information, to experimentally verify the necessity of annotating trigger information for \M{}. Category 1: 1) \textbf{DCFEE-O} \& \textbf{DCFEE-M}~\citep{yang-etal-2018-dcfee} proposed a key-event detection method to complete event arguments extraction by filling event table according to key-event mention and surrounding sentences. DCFEE-O only produces one event record from one key-event sentence. DCFEE-M tries to get multiple possible argument combinations. 2) \textbf{GreedyDec} \& \textbf{Doc2EDAG}~\citep{doc2edag} transformed the event extraction task into a directed cyclic graph path extension task based on entity information. GreedyDec only fills one event table entry greedily.  3) \textbf{DEPPN}~\citep{yang-etal-2021-document} used a multi-granularity non-autoregressive decoder to generate events in parallel based on document-aware representations. The second category comprises the current mainstream and advanced Event Extraction models to validate unique challenges of \M{} compared to other datasets. Category 2: 4) \textbf{BERT+CRF} is a sequence labeing model. 5) \textbf{EEQA}~\citep{du-cardie-2020-event} converted the event extraction task into the natural question answering task. 6) \textbf{TEXT2EVENT}~\citep{text2event} is a sequence-to-structure generation paradigm that can directly extract events from the text in an end-to-end manner. The third category consists of models that complete the Event Argument Extraction task based on golden triggers, in order to highlight the importance of resolving error propagation issues on \M{} in Event Extraction tasks. Category 3: 7) \textbf{PAIE}~\citep{ma-etal-2022-prompt} constructed prompts considering argument interaction to extract event arguments.

\paragraph{Evaluation} Following the widely-used setting, we report the micro Precision, Recall, and F-1 scores for event detection and event argument extraction as our evaluation metrics~\citep{MAVEN,tong-etal-2022-docee}.

\subsection{Overall Experimental Results}
Overall experimental results can be seen in Tabel \ref{overall-performance}. From these results we could find that: 1) Almost all models have relatively low performance on \M{}, which suggests that \M{} is challenging and that event extraction for the military domain deserves further exploration. 2) Models not using trigger information are significantly less effective than those using trigger information. Due to the specificity of the military domain, military events are always centered around several specific types of entities. In the event schema, the overlap of argument roles is higher compared to other domains, and it is difficult to accurately distinguish different events. Therefore, incorporating trigger information for event types is conducive to better accomplishing the military event extraction task. Among them, PAIE used golden trigger information, and the event argument extraction effect is much better than other models. 3) It is worth noting that BERT+CRF performs well on \M{} due to the advantage of CRF in modelling multi-event correlation, as can be seen from Figure \ref{fig.multi-evens}, \M{} has a high percentage of multi-events, and CRF is helpful in multi-event information extraction.

\begin{table*}
\centering
\setlength{\tabcolsep}{3.5mm}{
\begin{tabular}{lrrrrrr}
\toprule
\multirow{2}{*}{\textbf{Models}} & \multicolumn{3}{c}{\textbf{Event Detection}} & \multicolumn{3}{c}{\textbf{Event Argument Extraction}} \\
~& \textbf{Trg-P} & \textbf{Trg-R} & \textbf{Trg-F1} & \textbf{Arg-P} & \textbf{Arg-R} & \textbf{Arg-F1} \\
\midrule
DCFEE-O & - & - & - & 30.3 & 22.3 & 25.7 \\
DCFEE-M & - & - & - & 26.4 & 22.0 & 24.0 \\
GreedyDec & - & - & - & 39.4 & 19.9 & 26.4 \\
Doc2EDAG & - & - &  - & 54.3 & 23.9 & 33.2 \\
DEPPN & - & - & - & 38.2 & 35.0 & 36.5 \\
BERT+CRF & 73.1 & 77.7 & 75.3 & 63.1 & 52.3 & 57.2 \\
EEQA & 65.8 & 80.5 & 72.4 & 39.0 & 39.1 & 39.0 \\
TEXT2EVENT & 30.1 & 60.6 & 40.2 & 31.3 & 41.3 & 35.5 \\
PAIE & - & - & - & 72.0 & 67.0 & 69.4 \\
\bottomrule
\end{tabular}
}
\caption{\label{overall-performance}Overall event detection and event argument extraction performance on \M{}}
\end{table*}

\subsection{Analyzes on Data Size}
\begin{figure}[!ht]
\begin{center}
\includegraphics[scale=0.065,trim=5 80 2 35,clip]{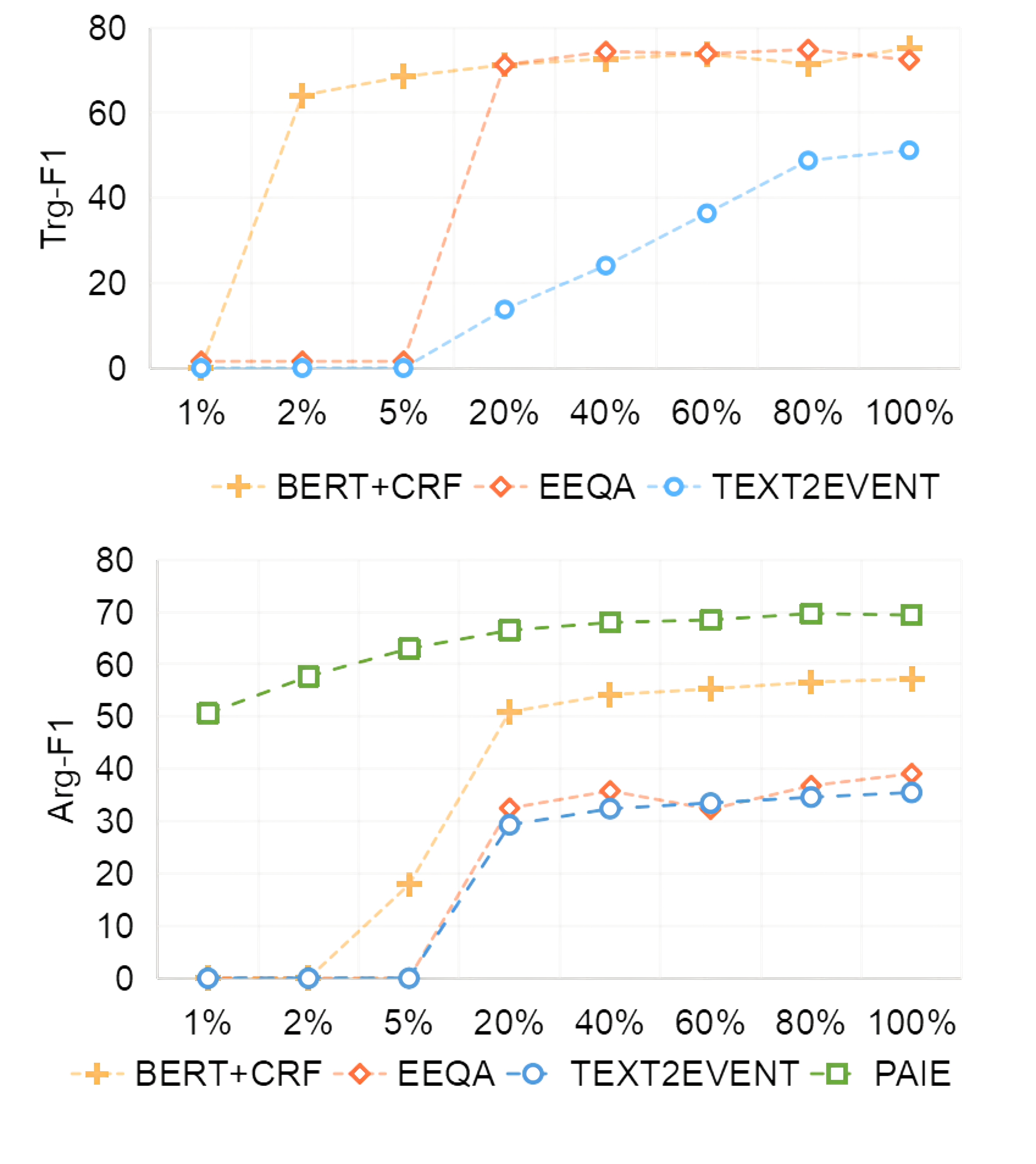} 
\caption{Model Performance~(Trg-F1 \& Arg-F1) change along with different training set data size.}
\label{fig.data-size}
\end{center}
\end{figure}

In this section, we selected several models with better performance to analyze the benefits of a larger data scale. We randomly choose different proportions of documents from the \M{} training set and compare the model performances (Trg-F1 for event detection and Arg-F1 for event argument extraction) trained with different data sizes. Results can be seen in Figure \ref{fig.data-size}. We can observe that \M{} can sufficiently train the models.

\subsection{Analyzes on Co-reference Arguments}
\label{sec:coref}
\begin{figure}[!ht]
\begin{center}
\includegraphics[scale=0.4,trim=5 10 2 55,clip]{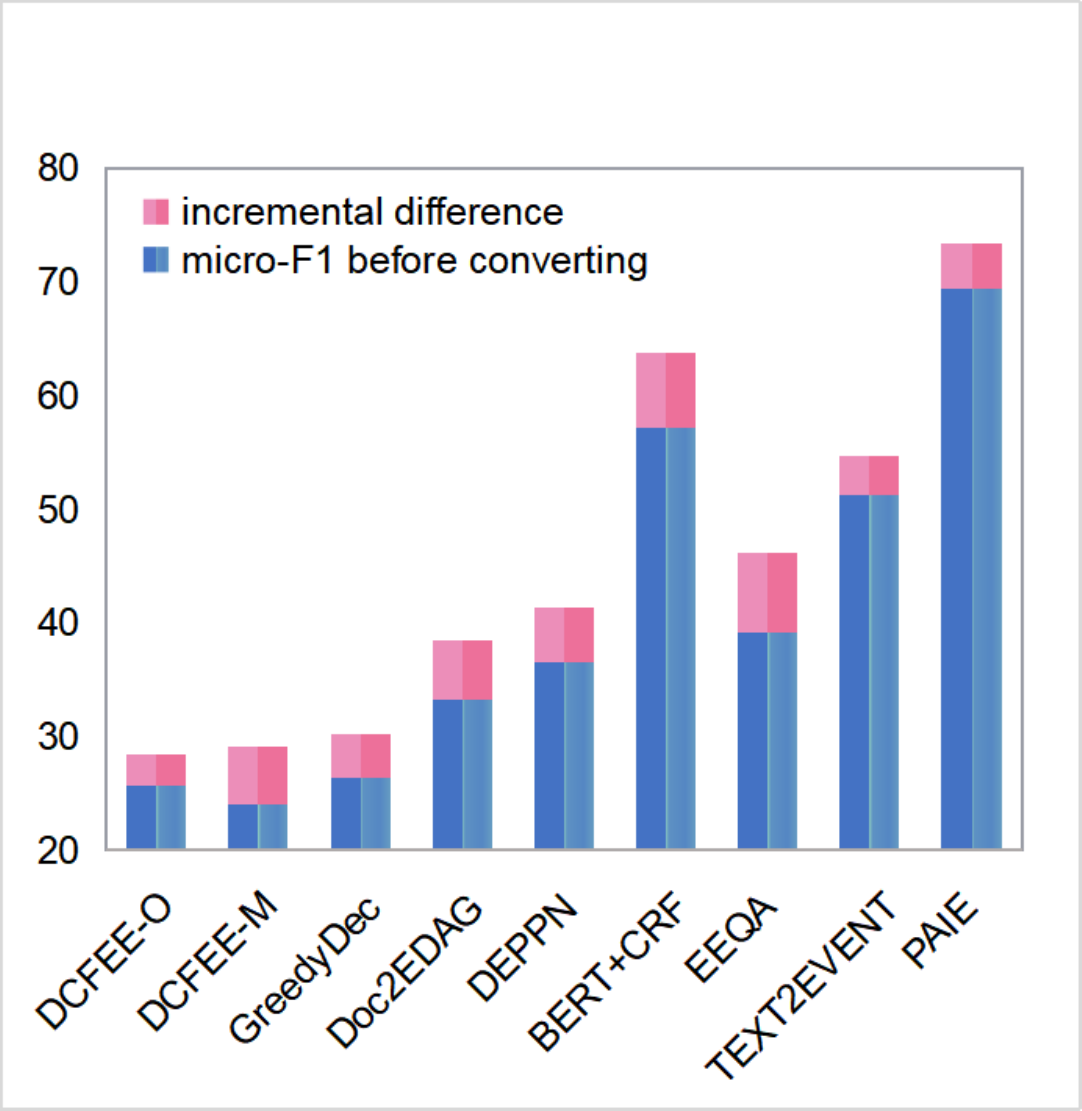} 
\caption{Converted metrics after considering co-reference arguments.}
\label{fig.convert-metrics}
\end{center}
\end{figure}

Since the same argument contained in the event in the document may appear several times with the same text and different offsets, or in the form of abbreviation, code name, etc., i.e., there exists co-reference arguments. The extraction of co-reference arguments can also express the event information. For example, in the sentence \emph{``The Seal nuclear submarine is a third-generation Russian submarine, construction of which began in 1991, and in 2008 the Seal underwent navigational tests in the Sea of Japan.''}, extracting ``The Seal nuclear submarine'' at the beginning of the sentence and ``the seal'' in the middle of the sentence both can be used as the Equipment argument of the Experiment event. Judging the extraction result only based on the events list does not meet the practical requirements. If co-reference argument information is taken into consideration, the performance of event argument extraction will be obviously improved as can be seen in Figure \ref{fig.convert-metrics}.

\subsection{Error Analysis}
In order to better understand the difficulties of \M{}, We randomly selected 50 samples to analyze errors that occurred in event detection and event argument extraction and categorized these errors into 3 main groups. 

\paragraph{Identification mistakes.} It is the most common error accounting for nearly 50\% that negative instances are misclassified to positive types (FP) or positive instances are misclassified to N/A (FN). This suggests that understanding event semantic information based on complex text still requires a great deal of effort.

\paragraph{Majority bias.} Due to the inherent unbalanced characteristics of \M{}, when dealing with complex texts, instance tends to be classified into types of higher frequency of occurrence. This error type accounts for about 20\% of all errors. Since we did not apply data augmentation or balancing during dataset construction and maintain the realworld distribution in \M{}, it is expected that models that can cope with this problem and accomplish the extraction of fewer sample events more efficiently will be devoloped.

\paragraph{Extraction boundary.} In this situation, extracted argument tokens are more or less than the actual labelled argument. For example, the correct equipment argument of Deploy is the \emph{SAD anti-missile system}, but the extracted argument is the localized text \emph{anti-missile system}, which omits key information. Another example is that location argument should be extracted is \emph{South Korea} rather than \emph{located in South Korea}, which introduces irrelevant information. How to better determine the extraction boundaries so that the extracted information is concise and effective deserves further research. This is one of the major reasons for the unsatisfactory performance of the event extraction model, with errors accounting for about 32\%.

\section{Related Work}
Since event extraction is one of the important tasks in natural language processing and is significant in practical applications, various datasets are constructed. As shown in Table \ref{compared-with-other-datasets}, the datasets constructed for different subtasks and different text granularity have their own characteristics. The most widely used dataset is ACE2005~\citep{ace2005}, which defines 8 event types, 33 event subtypes and 30 argument roles, and contains 599 documents and 4,480 events. MAVEN~\citep{MAVEN} is the largest event detection dataset, which defines 168 event types and labels 19,640 events. For the document-level event extraction task, the commonly used datasets are RAMS~\citep{ebner-etal-2020-multi} and WikiEvents~\citep{WIKIEVENTS}, with a smaller number of documents, 9124 and 246, respectively. ChfinAnn~\citep{doc2edag} is constructed using distant supervision to assist in the construction, with a sizable scale, but it does not contain event trigger information, and can only be used for the event argument extraction task.

Most of the current relevant datasets are oriented to general domains or financial domains. There are also some datasets oriented to other specific domains, such as biomedical domain~\citep{pyysalo-etal-2013-overview}, literary domain~\citep{sims-etal-2019-literary}, terrorist attack events~\citep{grishman-sundheim-1996-message}, breaking news~\citep{CEC}, etc. To the best of our knowledge, there are currently no publicly available English datasets specifically dedicated to event extraction tasks in the military domain. However, some existing datasets do include certain military events. For instance, the ACE2005 dataset contains the ``Conflict.Attack'' category, while the MAVEN dataset includes categories like ``Attack'' and ``Defending''. Additionally, the MUC-4 dataset encompasses information on terrorist attack events. The volume of military event-related data in these datasets is relatively limited and not easily distinguishable for specialized research purposes.

Recently, there have been some efforts using ChatGPT for data annotation in areas like Named Entity Recognition and Relation Extraction, as indicated in references~\citep{gp3-annotator,goel2023llms, zhang2023llmaaa}. However, due to the complex data structure involved, event extraction tasks prove more challenging for annotation. The potential reduction in manpower, time and costs were also limited. Considering \M{}'s status as a pioneering dataset in the field, we aimed for utmost authenticity, reliability, and quality, ultimately opting for manual annotation. Nevertheless, we admit that the use of new techniques remains a promising avenue for future research and deserves more attention.

\section{Conclusion}
To fill the gap between military event extraction and mainstream researches, we proposed a large-scale document-level event extraction dataset, \M{}, which was developed based on open-source Chinese military news. \M{} is an important complementary for existing event extraction datasets, and is helpful to advance the development of event extraction task research, particularly facilitate progress in the military domain. Compared to existing datasets, \M{} is oriented to the military domain, featuring a higher proportion of overlapping events and longer arguments due to the characteristics of military texts, thereby increasing the difficulty of extraction. Additionally, it includes annotations for co-reference arguments, making the assessment of extraction results more rational. Experiments demonstrate that \M{} is challenging and event extraction in military domain remains an open issue.

\section{Limitations}
We have to admit that \M{} has some limitations. 1) Limited event types and roles. Due to the confidentiality and sensitivity of the military domain, there is limited publicly available information, and the development of event schema and annotation criteria relies heavily on domain expert knowledge, which is not easily available. We finally defined 8 event types that are more common and currently suitable for public research. In order to improve the diversity, applicability, and impact of \M{}, we expected that more authoritative experts could be invited to increase and refine the event types. 2) Choose of language and annotation methodology. Due to the large size of \M{} and it is annotated based on documents, the annotated information is also relatively complex, in the existing attempts to annotate with the help of large language models, the results are not satisfactory and the cost that can be saved is very limited. As the first dataset in the military domain, we want to ensure the quality of the dataset as much as possible, and finally choose pure manual annotation. Although we designed a two-stage multi-turns annotation approach to improve the efficiency while guaranteeing the quality, the cost of time, manpower, and expense for the data annotation is still very high, so we only annotate the Chinese corpus at the moment. Because Chinese event extraction is also a relatively large and active community. But in the future, expanding \M{} to other languages and further try to annotate with new techniques are both necessary.

\section{Acknowledgements}
We thank the anonymous reviewers for their insightful comments and suggestions, researchers in the event extraction field for their excellent studies, data annotators for their hard work and all our team members for their kind support. We also would like to express special gratitude to Omnievent\footnote{\url{https://github.com/THU-KEG/OmniEvent}} \cite{peng2023omnievent}, a comprehensive, fair, and user-friendly toolkit for event understanding due to the substantial assistance throughout the evaluation of \M{}. In addition, because \M{} was used to support a competition\footnote{\url{https://www.datafountain.cn/competitions/987}} on the joint extraction of specific domain multi-events information held by \emph{Laboratary for Big Data and Decision}, \emph{National University of Defense Technology}, we also want to thank all experts and participants of this competition. 

This work was supported by National Science Foundation of China (Grant Nos.62376057) and the Start-up Research Fund of Southeast University (RF1028623234). 
All opinions are of the authors and do not reflect the view of sponsors.

\nocite{*}
\section{Bibliographical References}\label{sec:reference}

\bibliographystyle{lrec-coling2024-natbib}
\bibliography{lrec-coling2024-example}

\label{lr:ref}
\bibliographystylelanguageresource{lrec-coling2024-natbib}
\bibliographylanguageresource{languageresource}

\end{document}